%% file: main.tex
\documentclass[10pt,twocolumn,letterpaper]{article}

\usepackage[pagenumbers]{cvpr} 
\usepackage{algorithm}
\usepackage{algorithmic}
\usepackage{amsmath}
\usepackage{amssymb}
\usepackage{pifont}
\usepackage{multirow}
\usepackage{newfloat}
\usepackage{listings}
\usepackage{amssymb} 
\usepackage{amsmath}
\usepackage{makecell}
\input{preamble}
\definecolor{cvprblue}{rgb}{0.21,0.49,0.74}
\usepackage[pagebackref,breaklinks,colorlinks,allcolors=cvprblue]{hyperref}

\def\paperID{*****} 
\def\confName{CVPR}
\def\confYear{2026}

\title{Think, Then Verify: A Hypothesis–Verification Multi-Agent Framework for Long Video Understanding}

\author{
    Zheng Wang$^{1, 2}$\quad
    Haoran Chen$^{1}$\quad
    Haoxuan Qin$^{1}$\quad
    Zhipeng Wei$^{3}$\quad
    Tianwen Qian$^{4}$\quad
    Cong Bai$^{1, 2}$\thanks{Corresponding Author: Cong Bai.}
    \\
    \{zhengwang, 211124120004, 302023315145, congbai\}@zjut.edu.cn, \\
    zwei@icsi.berkeley.edu, twqian@cs.ecnu.edu.cn\\
    $^{1}$ College of Computer Science, Zhejiang University of Technology, Zhejiang, China \\
    $^{2}$ Zhejiang Key Laboratory of Visual Information Intelligent Processing, Zhejiang, China \\
    $^{3}$ UC Berkeley, CA, USA \\
    $^{4}$ College of Computer Science and Technology, East China Normal University, Shanghai, China
    \\
} 

\begin{document}
\maketitle
\input{sec/0_abstract}    
\input{sec/1_intro}

\input{sec/2_related}
\input{sec/3_method}
\input{sec/4_experiment}
\input{sec/5_conclusion}
{
    \small
    \bibliographystyle{ieeenat_fullname}
    \bibliography{main}
}

\input{sec/X_suppl}

\end{document}

%% file: preamble.tex
\usepackage{tcolorbox}

%% file: sec/0_abstract.tex
\begin{abstract}
Long video understanding is challenging due to dense visual redundancy, long-range temporal dependencies, and the tendency of chain-of-thought and retrieval-based agents to accumulate semantic drift and correlation-driven errors. 
We argue that long-video reasoning should begin not with reactive retrieval, but with deliberate task formulation: the model must first articulate what must be true in the video for each candidate answer to hold. This thinking-before-finding principle motivates VideoHV-Agent, a framework that reformulates video question answering as a structured hypothesis–verification process.
Based on video summaries, a Thinker rewrites answer candidates into testable hypotheses, a Judge derives a discriminative clue specifying what evidence must be checked, a Verifier grounds and tests the clue using localized, fine-grained video content, and an Answer agent integrates validated evidence to produce the final answer.
Experiments on three long-video understanding benchmarks show that VideoHV-Agent achieves state-of-the-art accuracy while providing enhanced interpretability, improved logical soundness, and lower computational cost. We make our code publicly available at: \url{https://github.com/Haorane/VideoHV-Agent}.
\end{abstract}

%% file: sec/1_intro.tex
\section{Introduction}
\label{sec:intro}

\begin{figure}[t]
    \centering
    \includegraphics[width=1.0\linewidth]{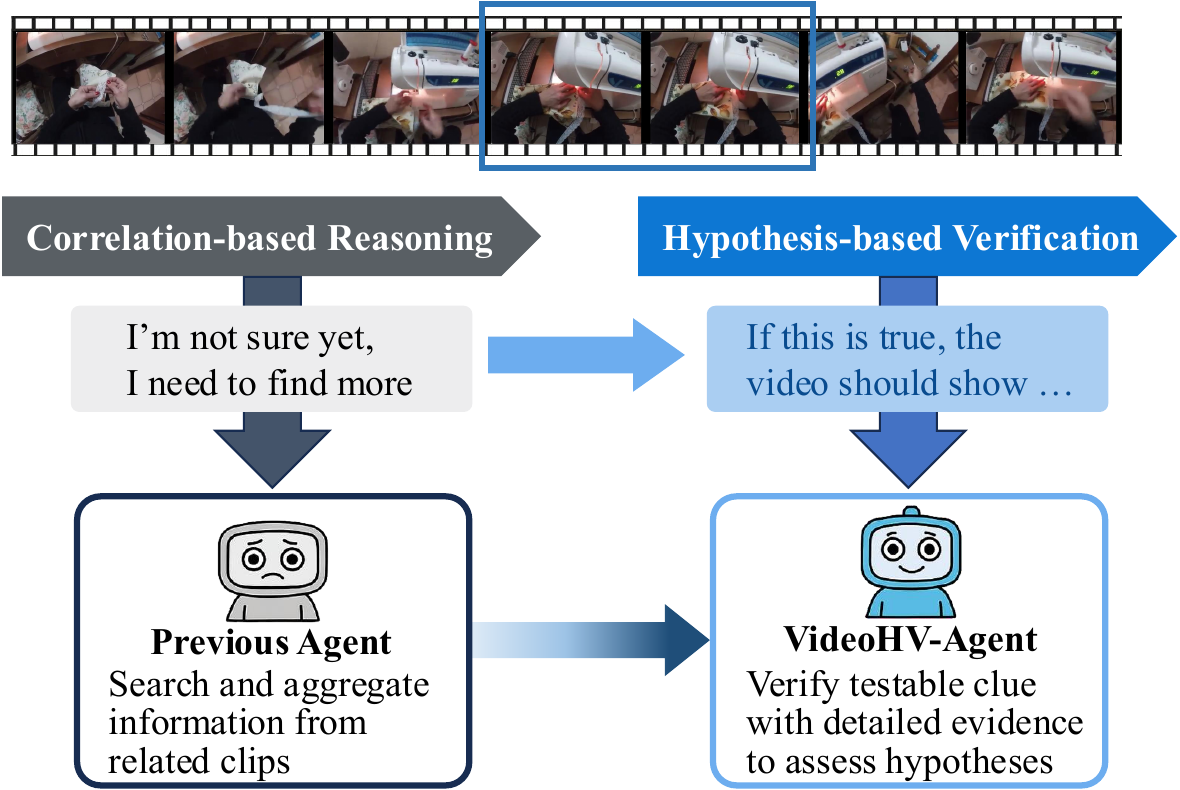}
    \caption{From correlation-based search to hypothesis verification: prior agents search and aggregate related clips, while VideoHV-Agent verifies testable clues with focused visual evidence.}
    \label{fig:cover}
\end{figure}
Large Language Models (LLMs) have advanced video understanding substantially, showing strong multimodal comprehension and emerging long-range reasoning capabilities.
Yet long-form video question answering (VideoQA) remains difficult: models must process dense, redundant content and reason over extended temporal spans, which often leads to information overload and cascading reasoning errors~\cite{zhang2024omagent}. 
Although chain-of-thought (CoT) prompting introduces stepwise reasoning~\cite{chen2024visual}, long reasoning chains are prone to semantic drift and error accumulation~\cite{cheng2024chainlm, wang2025streameqastreamingvideounderstanding}. Moreover, processing every frame is computationally prohibitive and leads to mixing of irrelevant content.

To cope with the extended temporal scale, mainstream approaches either (i) downscale inputs via key-frame/clip selection~\cite{zhang2024simple,park2024too}, or (ii) adopt multi-stage pipelines that first localize candidate moments and then reason on them~\cite{min2024morevqa,montes2025viqagent, li2025clivisunleashingcognitivemap}.
More recently, agent-based frameworks~\cite{fan2024videoagent, wang2025videotree, liao2024videoinsta} have emerged as a promising approach for long video understanding. These systems typically harness the observation, reasoning, and action capacities of LLMs. They use a pre-trained captioning model to obtain offline textual descriptions, then let agents plan multi-step information retrieval: iteratively search and aggregate video clips that appear semantically related to the question.

However, existing agent-based frameworks remain fundamentally \emph{correlation-driven}. On the one hand, most planners decompose the video’s complexity, such as video length, redundancy, and fine-grained information. While they overlook the question’s complexity, e.g., compositional constraints over multiple entities, temporal ordering, and causal preconditions,  each central to long-form VideoQA. As a result, long-range reasoning remains weak and early retrieval errors tend to propagate across subsequent steps~\cite{lin2025neighborretrbalancinghubcentrality,10.1145/3731715.3733330}. On the other hand, retrieval is treated as a reactive process. Agents repeatedly search for clips that correlate with the current plan, then re-plan based on whatever is found, leading to expensive trial-and-error cycles without explicitly checking whether the evidence actually supports or refutes a candidate answer.

Thus, the core difficulty of long-form VideoQA is not merely locating relevant clips but determining \emph{what should be looked for in the first place}. We argue that reasoning should begin with deliberate task formulation: before gathering evidence, the system must articulate what must be true in the video for each possible answer to hold. This ``thinking before finding'' principle reframes long-video reasoning as a structured inquiry rather than a purely correlation-driven search, as shown in Fig.~\ref{fig:cover}.

To this end, we propose \textbf{VideoHV-Agent}, a multi-agent framework that recasts long-video understanding as a \emph{hypothesis–verification} process, thus achieving ``thinking then verify'' principle.
After obtaining textual descriptions of sampled frames, we first build a query-conditioned video summary to quickly estimate which answers are plausible under coarse evidence. Then we can reason about which answers could be correct and why. Because the summary is inherently coarse, this preliminary reasoning is uncertain and must be turned into concrete information needs.
VideoHV-Agent addresses this via a two-stage reasoning pipeline. In Stage~1, a Thinker agent rewrites each answer option into an explicit, testable hypothesis that specifies key entities, actions, and temporal–causal constraints, while a Judge agent generates a concise clue that captures the minimal visual observation needed to distinguish among hypotheses. In Stage~2, a Verifier agent grounds this clue in the video, localizes a small set of relevant clips, and invokes fine-grained captioning to test whether the observed evidence supports, partially supports, or refutes the clue. Finally, an Answer agent integrates the verified evidence with the summarized context to answer along with a transparent reasoning chain.
By shifting from correlation-based retrieval to hypothesis–verification, VideoHV-Agent enables evidence-based, logically consistent, and interpretable reasoning for long-form VideoQA.
To summarize, our main contributions are:

\begin{itemize}
    \item We introduce a \textbf{hypothesis–verification} paradigm for long-form VideoQA, which reasons by first formulating testable hypotheses and then validating them against video evidence.
    
    \item We implement this paradigm as a multi-agent framework, \textbf{VideoHV-Agent}, where
    a Thinker agent that generates hypotheses and a Judge agent that distills a discriminative clue, 
    a Verifier that performs clue-guided evidence gathering and verification with detailed video analysis, 
    and an Answer agent that integrates validated evidence into a final decision.
    
    \item Extensive experiments on three long-video understanding benchmarks demonstrate that VideoHV-Agent achieves state-of-the-art performance with improved interpretability, logical accuracy, and computational efficiency.
\end{itemize}

%% file: sec/2_related.tex
\section{Related Work}
\subsection{Chain-of-Thought Reasoning}
Chain-of-Thought (CoT) prompting~\cite{wei2022chain} has been widely adopted to guide multi-step reasoning in long video understanding. It enables LLMs to decompose complex questions into intermediate steps, mirroring human problem-solving. ViQAgent~\cite{montes2025viqagent} combines CoT reasoning with visual grounding, using tracked objects to align steps with video content. TraveLER~\cite{shang2024traveler} employs a pipeline of agents to execute an ``ask-locate-evaluate-replan'' loop, where each sub-question is derived and answered sequentially. VideoINSTA~\cite{liao2024videoinsta} improves spatial-temporal reasoning ability by introducing event-based and context-based information with multi-round self-reflective reasoning mechanisms, improving confident self-evaluation. AoTD~\cite{shi2025enhancing} incorporates automatically generated Chain-of-Thoughts (CoTs) into the instruction-tuning process, showing the importance of constructing well-designed CoTs for instruction-tuning.  Despite these advances, CoT-based systems often remain fragile over long video contexts due to semantic drift, where early reasoning errors accumulate. Moreover, most lack explicit mechanisms to validate intermediate reasoning outcomes, making them vulnerable to hallucinations or propagation of incorrect assumptions.

\subsection{Efficient Contextualization for Long Video}
To mitigate the processing burden of long videos, several methods abstract or condense the video context before reasoning.
LVNet~\cite{park2024too} and VideoTree~\cite{wang2025videotree} reduce the input size through keyword-based frame selection and hierarchical clustering, respectively. VideoMindPlace~\cite{huang2025building} further organizes activity zones via layout mapping and hand-object interaction to improve spatial alignment. LLoVi~\cite{zhang2024simple} employs short-term visual captioners to summarize clips, which are later aggregated by LLMs for long-form understanding for better efficiency. VideoAgent~\cite{fan2024videoagent} constructs a memory bank of salient objects and events, enabling more targeted reasoning. 
Another direction for efficient long context processing is to incorporate retrieval augmented generation. Video-RAG~\cite{luo2024video} delivers auxiliary retrieval requests for OCR, ASR, and object information from the video, and the extracted auxiliary texts are combined with the query and the video to generate the response.
Grounding-based approaches, instead, try to narrow down the searching window during reasoning. TraveLER~\cite{shang2024traveler} employs grounding during sub-question answering by retrieving moments based on keyframe relevance. MoReVQA~\cite{min2024morevqa} proposes a multi-stage planning with grounding and reasoning abilities. Though these strategies enhance efficiency and visual alignment, they often rely on static pipelines and do not revisit raw evidence dynamically. Importantly, they lack logical validation across reasoning stages, limiting their reliability in multi-step inference settings.

\subsection{Specialized Agent Framework}
With the emergence of LVLMs, agent-based systems have become a prominent paradigm to improve modularity. For instance, VideoAgent~\cite{wang2024videoagent} builds an agent pipeline with the ability to predict answers, self-reflect, and find missing information. VideoAgent~\cite{fan2024videoagent} builds structured memory modules to store spatiotemporal information and support interactive planning across long sequences. 
MASR~\cite{cao2025masr} introduces a hierarchical, coarse-to-fine attention focusing mechanism in an agent-based framework.
OmAgent~\cite{zhang2024omagent} formulates LVU as a retrieval and generation process, and adopts a divide-and-conquer approach for task planning and execution.
ViQAgent~\cite{montes2025viqagent} combines the CoT framework with grounded reasoning powered by a specialized object grounding model.
VideoAgent2~\cite{zhi2025videoagent2} performs uncertainty-guided information retrieval, the retrieval plan is adjusted each round for collecting extra information with the help of specialized tool models.
These agent-based designs demonstrate the importance of separating planning, summarization, and action in long-form visual processing, laying the groundwork for modular reasoning across video timelines.

\subsection{Multi-Agent Collaboration Framework}
A more recent trend involves collaborative multi-agent frameworks for video understanding. TraveLER~\cite{shang2024traveler} exemplifies this by distributing across dedicated LLM agents. Similarly, VideoMultiAgents~\cite{kugo2025videomultiagents} introduces modality-specific agents, each focusing on a particular input type such as vision or language. VideoMind~\cite{liu2025videomind} adapts a single LVLM to four role-specific agents with LoRA fine-tuning~\cite {hu2022lora}. While these architectures begin to emulate collective intelligence, most implement static or loosely coupled pipelines without explicit validation across agents or dynamic feedback loops. They tend to partition tasks by modality or function, but fall short of using the reasoning ability to enable verifiable cooperation.

%% file: sec/3_method.tex
\section{Method}
\label{sec:method}
\begin{figure*}[t]
    \centering
    \includegraphics[width=1.0\linewidth]{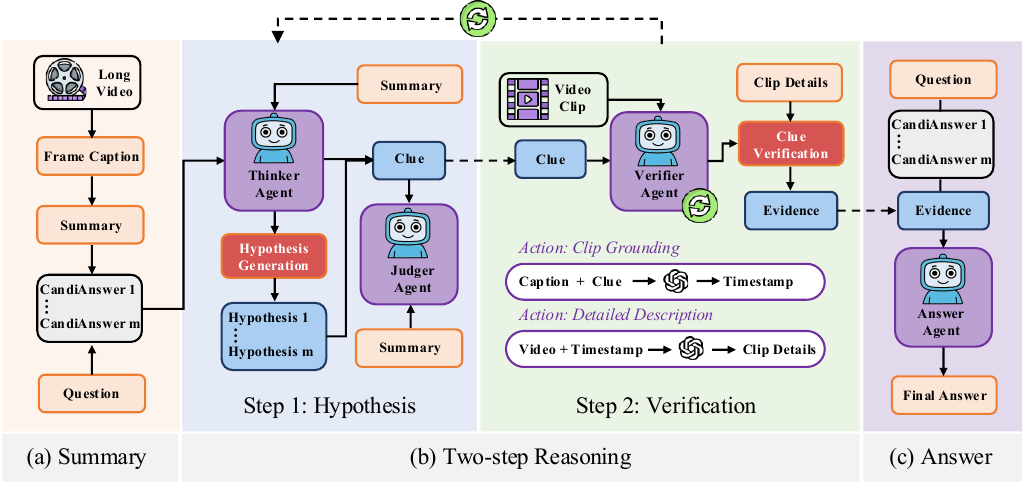}
    \caption{Overview of the proposed VideoHV-Agent framework. The framework first (a) summarizes the long video captions, then performs (b) two-step reasoning where a Thinker and a Judge agent rewrite options into hypotheses and a discriminative clue, and a Verifier agent grounds this clue to collect visual evidence, finally an Answer agent integrates the evidence to (c) answer the question.}
    \label{fig:framework}
\end{figure*}
\subsection{Problem setup}
Given a long video $V=\{f_{t}\}_{t=1}^{L}$ and a question $Q$ with multiple-choice options $O=\{o_i\}$, our goal is to return an answer $A$ supported by explicit, detailed video evidence $E$.

\subsection{Overall Framework}
We present \textbf{VideoHV-Agent}, a framework that consists of three stages as shown in Fig.~\ref{fig:framework}: \textbf{context summarization}, \textbf{two-step reasoning}, and \textbf{evidence integration}.
The two-step reasoning stage reformulates long video question answering as a \textit{hypothesis–verification} process, which can be iterated through a self-refinement loop. The pseudo code is presented in Alg.~\ref{alg:main}.
Specifically, VideoHV-Agent comprises mainly four cooperative agents. In the hypothesis generation step, a \textbf{Thinker} agent observes the summarized video description and proposes testable hypotheses $H$ for the candidate answers, while a \textbf{Judger} agent evaluates their quality and derives a concise clue $\kappa$ that specifies what needs to be verified. In the Verification step, a \textbf{Verifier} agent grounds this clue in the video, collecting visual evidence $E$ to test it; once the clue is verified (represented by the verification state $S$), an \textbf{Answer} agent combines the summary and the gathered evidence to produce the final answer $A$.
This framework enables interpretable, evidence-based reasoning over long video content.

\subsubsection{Context Summarization}
To address redundancy and complex temporal structure in long videos, we follow prior work~\cite{wang2024videoagent, zhi2025videoagent2} by first converting each frame into textual descriptions $P_{v}$ via captioning, and then deriving a compact, query-conditioned summary $P_{s}$ from the frame-level captions.  
Although the summarization step depends on the question and therefore cannot be performed fully offline, it is computationally lightweight compared to frame-level captioning, which requires repeated visual encoding.

In previous methods, frame captions and summaries are simply concatenated into a single long context~\cite{wang2024videoagent}, and the model consumes it for both local and global reasoning, incurring a time cost linear in the number of frames.  
In contrast, we decouple their roles: frame-level captions are used only for clip grounding, while the concise summary is used for global reasoning in other stages.  
This design preserves detailed information when necessary while keeping the overall context compact and efficient to process.

\subsubsection{Two-step Reasoning}
Given the summary, VideoHV-Agent can quickly narrow down plausible answers, but the summary alone is not sufficient to reliably resolve the question. In addition, directly reasoning over all frame-level captions is also impractical. It is time-consuming, and captions mainly describe salient content while often missing fine-grained relations or events needed for precise answering.
Therefore, VideoHV-Agent adopts a two-step hypothesis–verification process. 
First, it reasons about what information might be missing from the summary and formulates hypotheses that explicitly imagine the potentially unseen context. 
Then, it performs a verification step that checks whether detailed visual evidence satisfies these hypotheses, enabling accurate and efficient long-video reasoning.
Here we outline the two-step reasoning pipeline, with full methodological details provided in Sec.~\ref{sec:details}.

\noindent\textbf{Hypothesis Generation}. Given the summarized video context, the Thinker agent rewrites each answer candidate $o_i$ into a testable hypothesis $h_i$ that specifies what must be true in the video for $o_i$ to hold. Directly verifying all hypotheses one by one would ignore the logical relations among them. To address this, we introduce a Judge agent that evaluates the set of hypotheses and induces a discriminative clue $\kappa$, which condenses the key differences that need to be checked in order to distinguish among them.

\noindent\textbf{Hypothesis Verification}. Guided by clue $\kappa$, the Verifier grounds a minimal temporal context needed to evaluate it, to gather evidence to evaluate $\kappa$, invokes fine-grained tools (e.g., detailed captioning) to gather evidence, and outputs a structured status $\text{status}(\kappa) \in \{\text{VERIFIED}, \text{PARTIAL}, \text{NOT\_VERIFIED}\}$ together with a concise rationale.

\subsubsection{Self-Refinement Loop}
To improve robustness, VideoHV-Agent incorporates a self-refinement mechanism that mirrors human hypothesis revision.
When the verification status is inconclusive, we regenerate refined hypotheses and updated clues for an extra round of the reasoning stage.
Two regeneration prompts are used: (i) specificity enhancement, which makes hypotheses more concrete and testable when verification fails, and (ii) discriminability enhancement, which increases semantic contrast when hypotheses overlap.
This ensures that each reasoning loop progressively sharpens both the clarity of hypotheses and the precision of verification, yielding stable and logically grounded answers.

\subsubsection{Evidence Integration}
In the final stage, all verification results are integrated to infer the most plausible answer. With the summarized context and validated evidence, the video information is sufficient to resolve the question. The Answer agent re-evaluates each candidate option, checks for conflicts with the evidence, and constructs a reasoning chain outlining what was tested, observed, and supported or refuted. The final prediction is produced through explicit, evidence-grounded reasoning.

\subsection{Details of Two-step Reasoning}
\label{sec:details}
\subsubsection{Step 1: Hypothesis}
\noindent \textbf{Hypothesis Generation}.
We cast answer formation as hypothesis drafting: for each candidate option $o_{i}$, the Thinker Agent produces a testable hypothesis $h_{i}$ that, if observed in the video, would make $o_{i}$ correct.
The hypothesis $h_i$ specifies what must be true in the video for $o_i$ to hold, explicitly naming the salient entities/objects, actions/events, and temporal/causal constraints. 
Before generating hypotheses, the Thinker agent filters out ill-posed or clearly incorrect options using only the summarized context, thereby reducing irrelevant noise for the subsequent verification step and avoiding unnecessary reasoning cost. 
Formally, we map the option set $O=\{o_i\}$ to a hypothesis set $H=\{h_i\}$ with a one-to-one correspondence intended for later verification.

\noindent \textbf{Clue Generation}.
To enhance discriminability, the Judge agent further produces a concise \emph{clue} $\kappa$ for the hypothesis set $H$.
The clue summarizes the minimal observation that can distinguish $h_{i}$ from competing hypotheses, such as a specific object interaction, an event order, or a visual outcome that would hold only if $h_{i}$ were true.
This clue serves as focused guidance for the Verifier agent in the Verification step, defining what needs to be checked and what kind of evidence would support or refute the hypothesis.

\subsubsection{Step 2: Verification}
\noindent \textbf{Temporal Localization}.
Given the clue $\kappa$, the Verifier uses frame-level captions to localize the most probable temporal window where the clue appears, focusing on finding clip details on decisive evidence rather than the entire video.

\noindent \textbf{Detailed Captioning}.
After selecting the timestamp range, the Verifier revisits the raw frames within the window and invokes fine-grained captioning to extract detailed evidence for verification, ensuring that the system’s capabilities are not constrained by the initial visual to text translation. Each call processes at most five frames.

\noindent \textbf{Clue Verification}.
The clue verification status for $\kappa$ is one of
$\text{status}(\kappa) \in \{\text{VERIFIED}, \text{PARTIAL}, \text{NOT\_VERIFIED}\}$,
accompanied by a concise rationale (timestamps, entities, relations). 
\text{VERIFIED} means the clue can be verified by the evidence, and an answer can be raised upon this clue together with the summarized context. The Verifier agent then synthesizes all collected evidence into a reasoning trace that documents the logical inference, which is further used by the Answer agent.
\text{PARTIAL} indicates that part of the clue is verified by the observed, but requires additional evidence. Since a single round of evidence collection may be insufficient or error-prone, the Verifier can trigger additional rounds when the current assessment is inconclusive, explicitly specifying what further observations and frame ranges are needed. Then, VideoHV-Agent performs a small verification-only self-refinement loop to retrieve detailed descriptions from new timestamps and integrates them into the reasoning context. 
The \text{NOT\_VERIFIED} status indicates that the agent finds the clue sub-optimal and should be regenerated along with the hypothesis. When the NOT\_VERIFIED status is presented, the large hypothesis-verification self-refinement loop is activated.

\begin{algorithm}[t]
\caption{\textbf{VideoHV-Agent}}
\label{alg:main}
\begin{algorithmic}[1]
\REQUIRE video $V$; question $Q$; answer options $O$; frame-level caption $\mathcal{P}_v$; LLM $F_{llm}$; video summarizer $F_{vs}$; 
\ENSURE hypothesis $H$; clue $\kappa$; evidence $E$; verification status $S$; final answer $A$; 
\STATE \textbf{// Context Summarization}
\STATE $\mathcal{P}_f \gets F_{vc}(V)$
\STATE $\mathcal{P}_s \gets F_{vs}(\mathcal{P}_f, Q)$
\vspace{2pt}
\STATE \textbf{// Two-Stage Reasoning}
\FOR{t=1 \textbf{to} T}
\STATE \hspace{0.8em}\textbf{Step 1: Hypothesis Generation}
\STATE \hspace{1.6em}$H \gets F_{llm}(Q, O, \mathcal{P}_s, prompt_\text{Hypothesis})$
\STATE \hspace{1.6em}$\kappa \gets F_{llm}(H, \mathcal{P}_{s}, prompt_\text{Judge})$
\vspace{2pt}
\STATE \hspace{0.8em}\textbf{Step 2: Hypothesis Verification}
\STATE \hspace{1.6em}$E \gets F_{llm}(\kappa, \mathcal{P}_f, prompt_\text{Evidence})$
\STATE \hspace{1.6em}$S \gets F_{llm}(\kappa, E, prompt_\text{Verify})$
\STATE \hspace{1.6em}\textbf{if}
$S==\text{NOT\_VERIFIED}$ \textbf{then}
    \STATE \hspace{2.2em}  \textbf{continue}
    \STATE \hspace{1.4em} \textbf{else}
    \STATE \hspace{2.4em}\textbf{break}
\STATE \hspace{1.8em}\textbf{end if}

\ENDFOR
\vspace{2pt}
\STATE \textbf{// Evidence Integration}
\STATE $A \gets F_{llm}(Q, O, \mathcal{P}_s, \kappa, E, prompt_\text{Answer})$

\RETURN $A$
\end{algorithmic}
\end{algorithm}

%% file: sec/4_experiment.tex
\section{Experiment}
\subsection{Datasets and Metrics}
We compare VideoHV-Agent against strong zero-shot and supervised baselines under the accuracy metric, evaluating performance on three multi-choice video question answering benchmarks:
\noindent (i) EgoSchema~\cite{mangalam2023egoschema} is a large-scale test-only benchmark built on Ego4D, designed to evaluate deep reasoning over 3-minute egocentric videos. It contains 5,000 multiple-choice questions over 250+ hours of real-world footage, with 500 publicly labeled questions used for validation.
\noindent (ii) NextQA~\cite{xiao2021next} emphasizes causal and temporal reasoning in natural daily-life videos. It includes 5,440 videos and 48,000 questions, with 43s average length. We evaluate on the 570-video, 5,000-question validation split.
\noindent (iii) IntentQA~\cite{li2023intentqa} focuses on understanding character intent in narrative videos. It contains 4,303 videos and 16,000 QA pairs, with 567 videos and 2,134 questions in the test set. 
Results on longer videos, VideoMME-L~\cite{fu2025video}, are included in the supplementary materials.

\subsection{Implementation Details}
We extract all video frames at 1 fps for our experiments. Following the setup in~\cite{wang2024videoagent, zhi2025videoagent2}, we adopt LaViLa~\cite{zhao2023learning}, a clip-based captioning model, as the frame-level captioner for EgoSchema, and for NextQA and IntentQA, we use CogAgent~\cite{hong2024cogagent} to generate frame-level captions. GPT-4o~\cite{hurst2024gpt} is used as the LLM backbone for all four agents. Alternative backbone choices are detailed in the supplementary materials. For detailed captioning in the verification stage, we also employ GPT-4o to produce fine-grained textual descriptions. 

\begin{table}
  \caption{Zero-Shot Performance on EgoSchema subset compared to the state of the art.}
  \centering
  \begin{tabular}{lc}  
    \toprule
    \textbf{Method} & \textbf{Accuracy (\%)} \\ 
    \midrule
    TS-LLAVA~\cite{qu2024ts} & 57.8\\
    HCQA~\cite{zhang2024hcqaego4degoschema} & 58.8  \\
    VideoAgent~\cite{wang2024videoagent} & 60.2 \\
    VideoTree~\cite{wang2025videotree} & 66.2 \\
    LVNet~\cite{park2024too} & 68.2 \\
    Tarsier~\cite{wang2024tarsierrecipestrainingevaluating} & 68.6 \\
    LifelongMemory~\cite{wang2023lifelongmemory} & 72.0 \\
    VideoMultiAgents~\cite{kugo2025videomultiagents} & 75.4 \\
    VideoAgent2~\cite{zhi2025videoagent2} & 80.6 \\
    \midrule
    \textbf{VideoHV-Agent} & \textbf{81.0}  \\
    \bottomrule
  \end{tabular}
  \label{tab:egoschema}
\end{table}

\begin{table}[t]
    \centering
    \caption{Results on NextQA dataset compared to the state of the art.}
    \begin{tabular}{lccc}
        \toprule
       \textbf{Method} & \makecell[c]{\textbf{val set}} & \makecell[c]{\textbf{ATP-hard} \\ \textbf{subset}} \\
        \midrule
         \textit{Supervised} \\
        ViLA \cite{wang2024vila}        &74.4   & -     \\
        VideoChat2 \cite{li2024mvbench}  &79.5  & 68.2 \\
        LLaVA-OV \cite{li2024llava}     &  80.2 & -\\
        LinVT-Qwen2-VL \cite{gao2024linvt} &  \textbf{85.5} & 69.1 \\
        \midrule
        \textit{Zero-shot} \\
        SeViLA \cite{yu2024self}      & 63.6 & 50.8 \\
        VideoAgent \cite{wang2024videoagent}   & 71.3  & 58.4 \\
        TS-LLAVA~\cite{qu2024ts} & 73.6 & -\\
         LLoVi \cite{zhang2024simple}    &  73.8 & - \\
        VideoMultiAgents~\cite{kugo2025videomultiagents} & 79.6 & - \\
        VideoAgent2 \cite{zhi2025videoagent2}    &  80.5 & 68.2 \\
        \midrule
       \textbf{VideoHV-Agent} &  \textbf{80.7} &  \textbf{71.2} \\
        \bottomrule
    \end{tabular}
    \label{tab:exp_mainresult_nextqa}
\end{table}

\begin{table}[t]
  \caption{Zero-Shot performance on IntentQA compared to the state of the art. } 
  \centering
  \setlength{\tabcolsep}{0pt}
  \begin{tabular}{lc}
    \toprule
    \textbf{Method} & \textbf{Accuracy (\%)} \\
    \midrule
    SeViLA~\cite{yu2024self} & 60.9  \\
    IG-VLM~\cite{kim2024image} & 65.3  \\
    VideoTree~\cite{wang2025videotree} & 66.9  \\ 
    LLoVi~\cite{zhang2024simple} & 67.1 \\
    TS-LLAVA~\cite{qu2024ts} & 67.9  \\
    ENTER~\cite{ayyubi2025enter} & 71.5  \\
    LVNet~\cite{park2024too} & 71.7  \\
    VideoINSTA~\cite{liao2024videoinsta} & 72.8  \\
    VideoAgent2~\cite{zhi2025videoagent2} & 73.9  \\
    \midrule
    \textbf{VideoHV-Agent} & \textbf{75.6} \\
    \bottomrule
  \end{tabular}
  \label{tab:intentqa}
\end{table}

\begin{table}[t]
    \centering
    \caption{Ablation of framework components.}
    \begin{tabular}{lcc}
        \toprule
        \textbf{Method}  & \textbf{Accuracy(\%)} \\
        \midrule
        w/o hypothesis & 76.0 \\
        w/o clue & 78.6 \\
        w/o verification status & 74.0 \\
        \midrule
        \textbf{VideoHV-Agent} & \textbf{81.0} \\

        \bottomrule
    \end{tabular}
    \label{tab:hv_ablation}
\end{table}

\subsection{Main Result}
As shown in Table~\ref{tab:egoschema}, ~\ref{tab:exp_mainresult_nextqa} and ~\ref{tab:intentqa}, VideoHV-Agent achieved SOTA among zero-shot methods on all datasets.  It shows that video agents perform well on problems such as temporal order understanding, common sense understanding, and character behavior intention. For the NextQA ATP-hard subset, our accuracy improved significantly. VideoHV-Agent is not only effective at solving simple problems, but also performs even better on difficult ones. This demonstrates that VideoHV-Agent has the ability to solve complex problems. This is also consistent with our hypothesis verification method, which focuses more on causal reasoning. VideoHV-Agent shows outstanding performance on both splits, highlighting the robustness of the method.

\subsection{Ablation Study}
\noindent \textbf{Ablation of framework components.} As reported in Table~\ref{tab:hv_ablation}, we conduct comprehensive ablation studies to validate the effectiveness of each component in the proposed framework:

\noindent (i) \textbf{Without hypothesis generation.} Removing the hypothesis generation module and deriving clues directly from raw option differences reduces accuracy by 5\%. This indicates that explicit hypotheses, which capture key events, entities, and temporal–causal relations, provide crucial structure for downstream reasoning; without them, information loss leads to degraded performance.

\noindent (ii) \textbf{Without clue generation.} Disabling clue generation yields 78.6\% accuracy. Clues translate high-level hypotheses into concrete, verifiable visual evidence (e.g., behaviors, locations, temporal order, object interactions). Removing them weakens verification focus and reduces the effectiveness of targeted evidence collection.

\noindent (iii) \textbf{Without verification status.} In particular, removing the explicit verification status leads to a 7\% performance drop, demonstrating that verification is functionally necessary rather than a cosmetic explanation. This mechanism enables adaptive self-refinement, collects additional evidence when needed, and regenerates poor hypotheses or clues. Without it, refinement is disabled and performance drops substantially.

\noindent (iv) \textbf{Full model.} The complete framework with Thinker, Judger, Verifier, and Answer agents achieves the highest accuracy of 81.0\%.

\begin{figure}[t]
    \centering
    \includegraphics[width=1\linewidth]{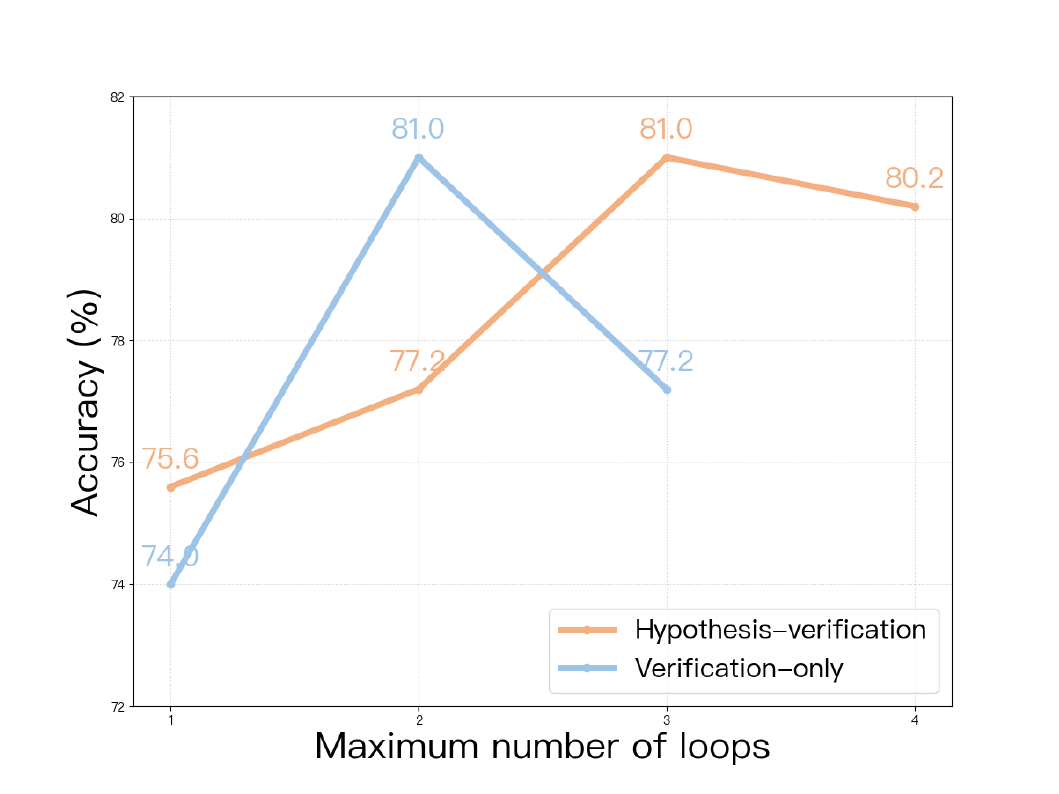}
    \caption{Ablations of different maximum number of loops.}
    \label{fig:max_iter}
\end{figure}

\begin{figure}[t]
    \centering
    \includegraphics[width=1\linewidth]{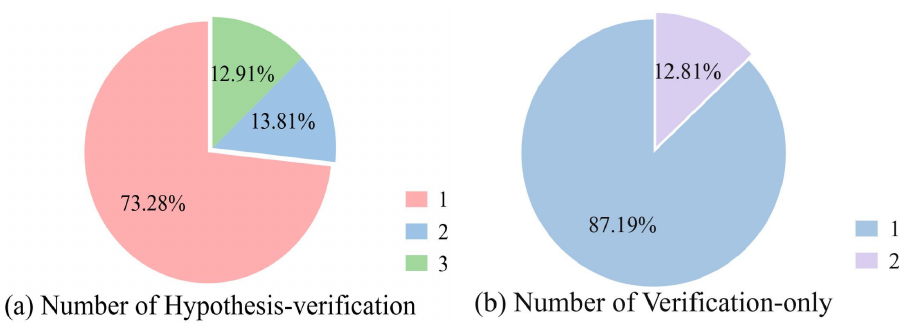}
    \caption{Proportion of samples with different numbers of loops.}
    \label{fig:real_iter}
\end{figure}

\noindent \textbf{Ablation of the number of loops.}
We investigate the effect of different maximum loop numbers on VideoHV-Agent. Hypothesis-verification refers to the number of self-refinement loop. 
Verification-only indicates the number of times the Verifier looks for additional information when verifying a hypothesis. As shown in Figure~\ref{fig:max_iter}, performance on the EgoSchema subset increases with more loops, reaches a peak, and then stabilizes. In Figure~\ref{fig:real_iter}, we show the actual number of loops performed by VideoHV-Agent. The largest proportion of loops is 1.
These results indicate that VideoHV-Agent typically identifies key information efficiently, while additional loops can further refine retrieval when necessary. Beyond a certain point, however, extra iterations provide marginal gains.

\subsection{Operational Efficiency}
Most agent-based VideoQA systems operate in an ``explore–reason–explore again'' fashion. The LLM agent repeatedly plans, retrieves new clips, updates its belief state, and then issues further queries. Each cycle triggers additional visual encoding, captioning, and long context reasoning, and the number of such cycles typically grows with video length. As a result, inference cost scales almost linearly with both the number of frames and the depth of the interaction loop, making these methods slow and sensitive to redundant content in long videos.

\begin{table}[t]
    \centering
    \caption{Averaged time cost comparison with other agent-based methods.}
    \begin{tabular}{lccc}
        \toprule
       \textbf{Method}  & \makecell[c]{\textbf{Time Cost(s)}} & \makecell[c]{\textbf{Accuracy(\%)}} \\
        \midrule
          VideoAgent~\cite{wang2024videoagent}       & 129.46 & 60.2   \\ 
         VideoTree~\cite{wang2025videotree}   & 160.21 & 66.2 \\
         VideoMultiAgents~\cite{kugo2025videomultiagents}  & 134.90 & 75.4 \\
        \midrule
        \textbf{VideoHV-Agent}   & \textbf{123.66} & \textbf{81.0} \\
        \bottomrule
    \end{tabular}
    \label{tab:time}
\end{table}

As shown in Table~\ref{tab:time}, VideoHV-Agent not only achieves higher accuracy than existing methods, but also offers substantially lower per-question latency, yielding a dual gain in performance and efficiency. 
This improvement comes from its design: frame captions are first condensed into a global summary, and the hypothesis-verification stages are guided by a discriminative clue that narrows the search to a minimal temporal window. The Verifier then analyzes only a small set of targeted frames with fine-grained captioning, rather than repeatedly scanning the entire video. By preserving decision relevant information while avoiding redundant frame-by-frame or multi-round inference, VideoHV-Agent maintains a relatively low overall reasoning time.

\begin{figure*}[t]
    \centering
    \includegraphics[width=1\textwidth]{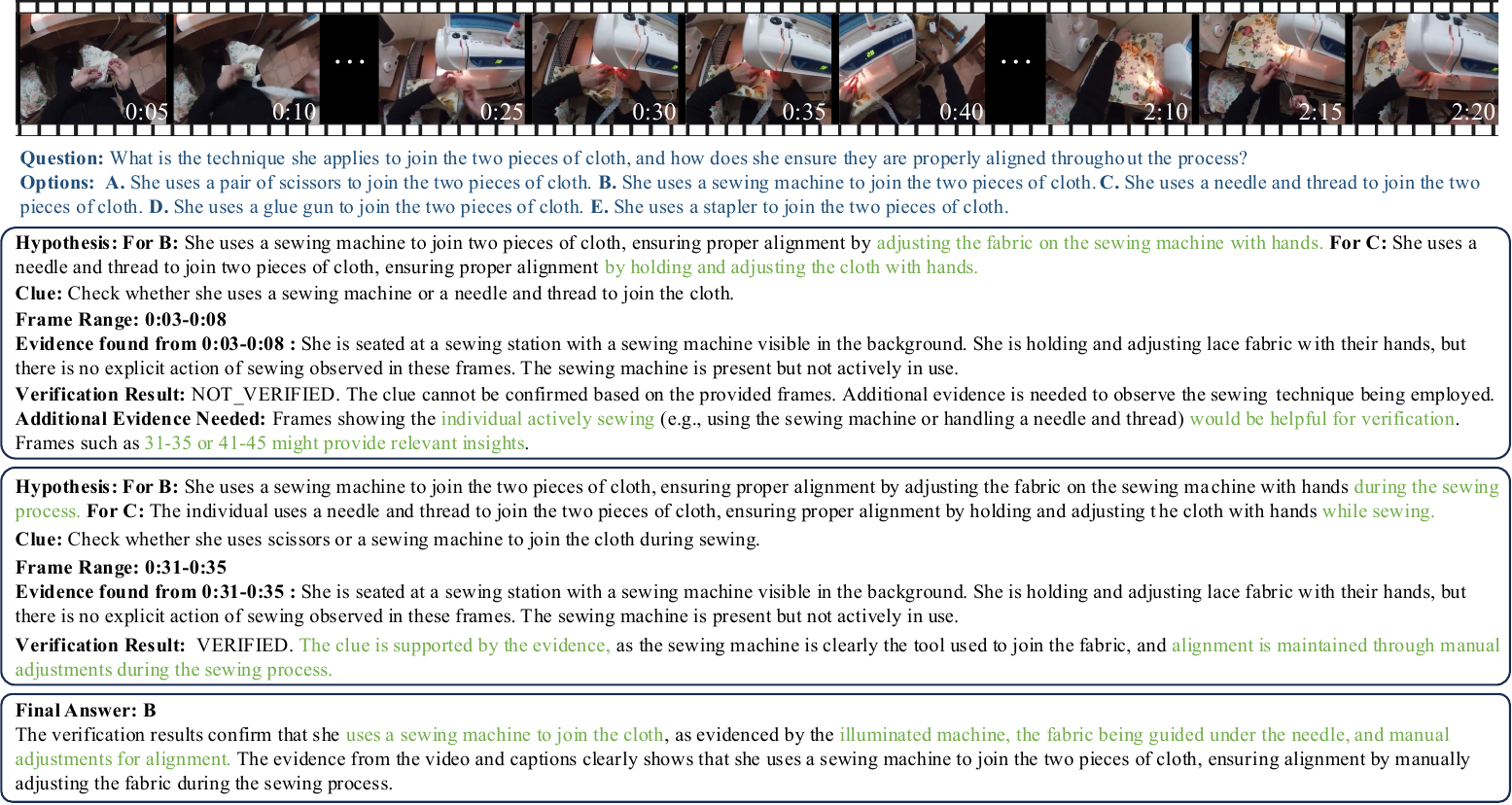}
    \caption{Qualitative study of event understanding in long videos. VideoHV-Agent uses hypothesis–verification to locate decisive evidence, highlighting its ability to avoid search purposefully and ground conclusions in explicit visual proof.}
    \label{fig:case}
\end{figure*}

\subsection{Performance Analysis by Question Type}
\begin{figure}[t]
    \centering
    \includegraphics[width=1\linewidth]{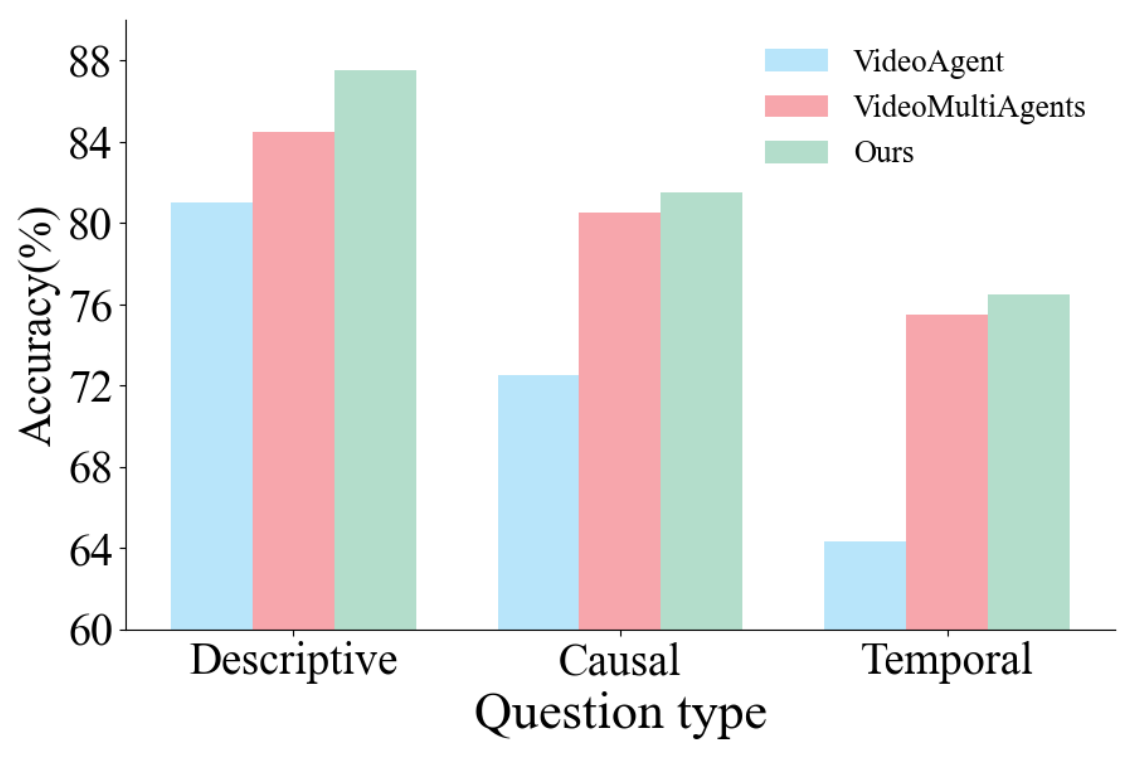}
    \caption{Comparison for different types of NextQA questions.}
    \label{fig:question_type}
\end{figure}

To demonstrate the effectiveness of our hypothesis-verification method, we analyze its accuracy across different question types on the NextQA dataset: (i) Causal, which examines cause–effect relationships; (ii) Temporal, which focuses on event order and interactions over time; and (iii) Descriptive, which emphasizes key scenes, objects, and actions.
As shown in Fig.~\ref{fig:question_type}, we compare our approach with VideoAgent and VideoMultiAgents. Our method achieves the best performance across all three categories within the agent-based video understanding framework, demonstrating the general effectiveness of the ``hypothesis-and-verification'' paradigm for commonsense, causal, and temporal reasoning.

\subsection{Qualitative  Study}
In Fig.~\ref{fig:case}, we qualitatively demonstrate the effectiveness of VideoHV-Agent. Given multiple possible cloth joint ways, the Thinker formulates testable hypotheses and specifies the visual clue needed to distinguish them. When inspecting early frames (3–8), the Verifier finds insufficient evidence and marks the clue as unverified rather than guessing. In a second round, it examines later frames (31–35), detects clear sewing actions (e.g., illuminated machine, fabric under the presser foot, manual alignment), and reaches a verified judgment consistent with the ground truth. This case demonstrates targeted evidence retrieval, robust uncertainty handling, and transparent reasoning that directly ties visual observations to logical decisions.

%% file: sec/5_conclusion.tex
\section{Conclusion}
We propose VideoHV-Agent, a multi-agent framework for long-form VideoQA built on a hypothesis–verification paradigm. Unlike prior single-agent or retrieval-based methods, it decomposes reasoning into explicit, verifiable steps executed by specialized agents.
By testing structured hypotheses instead of aggregating noisy evidence, VideoHV-Agent filters spurious information, focuses on decision-relevant observations, and enables transparent agent coordination. This results in more robust, interpretable, and logically consistent video understanding.
Experiments on EgoSchema, NextQA, and IntentQA show that VideoHV-Agent achieves state-of-the-art accuracy with strong efficiency and interpretability.
\section{Acknowledgements}
This work is partially supported by Natural Science Foundation of China (No. 62302453), Zhejiang Provincial Natural Science Foundation of China (No. LMS25F020003), and Shanghai Municipal Science and Technology Major Project 2025SHZDZX025G16.

%% file: sec/X_suppl.tex

\clearpage
\appendix

\renewcommand{\thetable}{S\arabic{table}}
\renewcommand{\thefigure}{S\arabic{figure}}

\setcounter{page}{1}
\setcounter{section}{0}
\setcounter{figure}{0}
\setcounter{table}{0}
\setcounter{equation}{0}
\maketitlesupplementary

\section{Result on VideoMME-L}
To further evaluate the generalization capability of our framework under broader video distributions and more diverse question types, we extend our experiments to VideoMME-L~\cite{fu2025video}, a large-scale benchmark designed for general VideoQA. Compared with EgoSchema, IntentQA, and NextQA, VideoMME-L covers more open-domain content and varied reasoning scenarios, providing a more comprehensive evaluation setting.

Table~\ref{tab:videomme-l} reports results on VideoMME-L, where our method remains strong, outperforming the recent Video Curious Agent (VCA) in ICCV’25~\cite{yang2025vca}. Under the same LLM/caption backbone (Table~\ref{tab:llm/cap}), we consistently outperform the standard chain-of-thought reasoning and achieve the largest gains over prior approaches.

In addition to accuracy, our framework demonstrates favorable efficiency across datasets. On NextQA, where the average video duration is 39.5s, our framework requires 74.48s per sample on average. On EgoSchema, with an average duration of 180s, the runtime is 123.67s. Notably, on the long-video benchmark VideoMME-L, which has an average duration of 2466.7s, our method only takes 181.82s per sample on average. These results indicate that our approach maintains stable and scalable computational cost even as video length increases substantially, highlighting its practicality for real-world long-form VideoQA scenarios.
\section{Results on Different Backbones}
To better understand the contribution of our hypothesis–verification architecture independent of backbone strength, we conduct additional controlled experiments with different LLM and caption backbones. Since existing methods adopt a variety of backbone configurations, directly comparing overall performance may obscure the architectural contribution. Therefore, we explicitly disentangle backbone effects by evaluating different backbone combinations within a unified setting.

In Table \ref{tab:egoschema}, 7 of the 9 baselines already use strong LLM backbones: 2 use GPT-4 and 5 use GPT-4o, while only TS-LLAVA and Tarsier do not.
To further isolate the contribution of the hypothesis–verification design, Table~\ref{tab:llm/cap} reports a fair backbone-controlled setting and also shows the baselines that do not perform hypothesis verification, but use standard chain-of-thought reasoning (CoT). Under identical backbone and captioner, our method achieves the best performance. Our framework remains consistently effective, indicating that our gains are not solely driven by using a stronger model.

\begin{table}[t]
\centering
\small
\caption{Results on VideoMME-L, with \textbf{VCA (ICCV’25)}.}
\label{tab:videomme-l}
\begin{tabular}{lcc}
\toprule
LLM & Methods & Accuracy (\%) \\
\midrule
\multirow{4}{*}{GPT-4o} 
 & CoT & 46.7 \\
 & VideoTree & 54.2 (+7.5\%) \\
 & VCA & 56.3 (+9.6\%) \\
 & \textbf{Ours} & \textbf{60.6 (+13.9\%)} \\
\bottomrule
\end{tabular}
\end{table}

\begin{table}[t]
  \caption{Extra EgoSchema results under different LLM/captioner.}
  \centering
  \resizebox{\columnwidth}{!}{
    \begin{tabular}{lccc}
      \toprule
      \textbf{LLM} & \textbf{Caption Model} & \textbf{Methods} & \textbf{Egoschema} \\
      \midrule
      \multirow{3}{*}{Claude-3-Haiku} 
        & - & CoT & 60.9 \\
        & LaViLa & LifelongMemory & 64.8 (+3.9\%) \\
        & LaViLa & \textbf{Ours} & \textbf{65.4 (+4.5\%)} \\
      \midrule
      \multirow{4}{*}{GPT-3.5}
        & - & CoT & 60.4 \\
        & GPT-4o & LVNet & 61.0 (+0.6\%) \\
        & LaViLa & LifelongMemory & 64.0 (+3.6\%) \\
        & LaViLa & \textbf{Ours} & \textbf{76.2 (+15.8\%)} \\
      \midrule
      \multirow{4}{*}{GPT-4o}
        & - & CoT & 66.0 \\
        & GPT-4o & LVNet & 68.2 (+2.2\%) \\
        & GPT-4o & LifelongMemory & 72.0 (+6.0\%) \\    
        & GPT-4o & \textbf{Ours} & \textbf{81.0 (+15.0\%)} \\
      \bottomrule
    \end{tabular}
  }
  \label{tab:llm/cap}
\end{table}

\section{Details of prompt in VideoHV-Agent}
We present all prompts used in the VideoHIV-Agent. First, frame-level information is integrated through Context Summarization. In the Hypothesis Generation stage, the agent observes the summarized video description and proposes a set of testable hypotheses for the candidate answers. In the Clue Generation stage, the agent evaluates the quality of these hypotheses and derives a concise clue that specifies what needs to be verified. In the Clue Verification stage, the agent grounds the clue in the video content and collects visual evidence to test it. When the verification result is inconclusive or lacks sufficient discriminative power, a hypothesis regeneration process is triggered. Specifically, Discriminability Enhancement addresses the problem of insufficient discrimination between hypotheses, while Specificity Enhancement makes the hypotheses more concrete and testable when verification fails. The final stage, Evidence Integration, integrates all verification results to infer the most plausible answer.

\begin{tcolorbox}[title=\textbf{Context Summarization},colback=SeaGreen!10!CornflowerBlue!10,colframe=RoyalPurple!55!Aquamarine!100!]
You are given some language descriptions of a first-person view video. The video is \{length\} seconds long. Each sentence describes a 1.0s clip. The descriptions are sequential and non-overlapping which cover the whole video exactly. Here are the descriptions: \{interval text\}. Please give me a \{words\} words summary. When doing summarization, remember that your summary will be used to answer this multiple choice question: \{question\}
\end{tcolorbox}

\begin{tcolorbox}[title=\textbf{Hypothesis Generation},colback=SeaGreen!10!CornflowerBlue!10,colframe=RoyalPurple!55!Aquamarine!100!]
[Instruction]\\
You are a reasoning planner. Given a video-related question and several answer options. Pay special attention to, if the option is clearly inappropriate for the context, do not generate a corresponding hypothesis. Rewrite remaining option into a testable hypothesis that can be verified from the video evidence.\\
Each hypothesis must specify:\\
- The key entities or objects involved,\\
- The main action or event,\\
- The temporal or causal relation among them.\\

[Inputs]\\
Question: \{question\}\\
Options: \{options\}\\
Context summary: \{video summary\}\\

[Output format]\\
Return option (It must be a complete option, not just numbers) and hypotheses.

\end{tcolorbox}

\begin{tcolorbox}[title=\textbf{Clue Generation},colback=SeaGreen!10!CornflowerBlue!10,colframe=RoyalPurple!55!Aquamarine!100!]
[Instruction]\\
You are a reasoning analyst.
Given several rewritten hypotheses corresponding to different answer options of a video question, analyze how distinct they are in terms of testable evidence in the video.
Your task has three parts:\\

Compare Hypotheses:\\
Identify the core difference in entities, actions, events, causal/temporal relations, or visual evidence type (e.g., spatial layout, sequence of actions, emotional expression).\\

\end{tcolorbox}

\begin{tcolorbox}[colback=SeaGreen!10!CornflowerBlue!10,colframe=RoyalPurple!55!Aquamarine!100!]

Generate a Distinguishing Clue:\\
Produce a concise description of what kind of video evidence could distinguish between these hypotheses.
For example: “Check whether the person hands the object before or after speaking,” or “Verify if the dog appears indoors or outdoors.”\\

Assign a Distinction Score (0–1):\\
Give a numeric score representing how distinguishable the hypotheses are based on likely video evidence, and provide the reasons for the score:
0.0–0.3: Hypotheses are too similar or overlapping\\
0.4–0.6: Moderate distinction but may require nuanced understanding\\
0.7–1.0: Strongly distinct, clearly testable difference\\
If the distinction $score < 0.5$, subsequent hypotheses need to be regenerated.\\

[Inputs]\\
Question: \{question\}\\
Hypotheses: \{hypotheses\}\\
Context summary: \{video summary\}
\end{tcolorbox}

\begin{tcolorbox}[title=\textbf{Clue Verification},colback=SeaGreen!10!CornflowerBlue!10,colframe=RoyalPurple!55!Aquamarine!100!]

[Instruction]\\
You are a reasoning verifier.
You will verify whether a distinguishing clue derived from multiple hypotheses can be supported or refuted by the provided video context information.\\
Follow this reasoning plan:\\
1. Clue Understanding\\
    Reinterpret the clue in plain terms: what needs to be verified?
    Identify what kind of evidence (objects, actions, relations, timing, spatial layout) would support or contradict it.

2. Contextual Search\\
    Examine the given video summary, transcript, or extracted description.
    Find sentences, events, or visual cues that are relevant to the clue.

3. Reasoning Trace\\
    Step by step, explain how the found evidence relates to the clue.
    Explicitly note whether the evidence supports or contradicts each hypothesis.
    Maintain logical transparency: show what was observed, what inference was drawn, and what conclusion followed.
    
\end{tcolorbox}

\begin{tcolorbox}[colback=SeaGreen!10!CornflowerBlue!10,colframe=RoyalPurple!55!Aquamarine!100!]
4. Final Output\\
    Summarize whether the clue was verified, partially verified, or not found.
    Provide a short reasoning trace and the relevant evidence snippet.\\

[Inputs]\\
Question: \{question\}\\
Clue: \{clue\}\\
Frame level caption: \{frame caption\}\\

[Output format]\\
``clue\_understanding'': ``Describe what is being tested'',\\
``evidence\_found": ``Summarize key details from the context or from tool retrieval'',\\
``reasoning\_trace": ``[\\
  Step 1: Identify action/event ...,\\
  Step 2: Compare with clue condition ...,\\
  Step 3: Draw inference ...\\
]'',\\
``verification\_result'': ``verified / partially\_verified / not\_verified'',\\
If you cannot verify clues well and which frames might be helpful for verification, describe what additional evidence is needed.
Call only one tool at a time and frame\_range no more than 5 frames.
If the verification result is not\_verified, it means the clue is not good and needs to be regenerated.
\end{tcolorbox}

\begin{tcolorbox}[title=\textbf{ Evidence Integration},colback=SeaGreen!10!CornflowerBlue!10,colframe=RoyalPurple!55!Aquamarine!100!]
[Instruction]\\
You are an answer reasoning agent.
Given a video question, and the verification results of the distinguishing clues, infer which option is best supported by the evidence.
You must:\\
1. Integrate Verification Results\\
2. Resolve Conflicts\\
    If multiple hypotheses are partially verified, reason which one is more strongly aligned with the overall context and clues. If all clues are unverified, indicate uncertainty and suggest that additional evidence is needed.\\
3. Generate a Transparent Reasoning Chain\\
    Summarize what was tested, what was found, and how it leads to your conclusion. Avoid just “guessing.” Show cause–effect reasoning.\\
4. Output a Final Answer\\
    Specify the chosen option (0/1/2/...) and briefly justify it based on evidence.\\
\end{tcolorbox}

\begin{tcolorbox}[colback=SeaGreen!10!CornflowerBlue!10,colframe=RoyalPurple!55!Aquamarine!100!]
5. Note that you should only refer to the Verification Results, not accept them wholesale.\\

[Inputs]\\
Question: \{question\}\\
Option: \{options\}\\
Clue Verification Results: \{verification\_outputs\}\\
Context summary: \{video summary\}\\

[Output format]\\
``reasoning\_summary'': ``Summarize the verification results.'',\\
``conflict\_resolution'': ``If any conflicting evidence exists, explain how it was resolved.'',\\
``final\_answer'': ``Choose from the options to return only numbers.'',\\
``explanation'': ``Give a concise, human-readable rationale connecting evidence to conclusion.''
\end{tcolorbox}

\begin{tcolorbox}[title=\textbf{Discriminability Enhancement},colback=SeaGreen!10!CornflowerBlue!10,colframe=RoyalPurple!55!Aquamarine!100!]
[Instruction]\\
You are a reasoning planner for video question answering.
The previous round of hypotheses had low distinction, meaning the hypotheses were too similar or not easily distinguishable based on video evidence.
Your task is to regenerate a new set of hypotheses—each still grounded in its respective option, but now rewritten to maximize their semantic and evidential differences so that they can be clearly tested and compared from the video.
Follow these reasoning steps:\\
1. Analyze Distinction Feedback\\
    Identify why the distinction score was low (e.g., overlapping actions, vague entities, lack of temporal or causal contrast).
    Determine what types of features (action, timing, spatial layout, emotion, interaction) can be emphasized to make them more distinct.\\
2. Regenerate Option-based Hypotheses\\
    Keep each hypothesis aligned with its original option’s meaning.
    Rewrite it to highlight unique, observable, and discriminative aspects of the event.
    Ensure that each hypothesis involves different entities, actions, outcomes, or temporal relations whenever possible.\\

\end{tcolorbox}

\begin{tcolorbox}[colback=SeaGreen!10!CornflowerBlue!10,colframe=RoyalPurple!55!Aquamarine!100!]
[Inputs]\\
Question: \{question\}\\
Options: \{option\}\\
Previous hypotheses: \{previous\_hypotheses\}\\
Verification feedback: \{verification\_feedback\}\\
Context summary: \{video summary\}\\

[Output format]\\
Return option (It must be a complete option, not just numbers) and hypothesis.
\end{tcolorbox}

\begin{tcolorbox}[title=\textbf{Specificity Enhancement},colback=SeaGreen!10!CornflowerBlue!10,colframe=RoyalPurple!55!Aquamarine!100!]
[Instruction]\\
You are a reasoning planner improving video-based hypotheses for a multiple-choice question.
Each hypothesis must correspond directly to one answer option.
The previous verification phase failed to confirm the clues, so you need to regenerate hypotheses that are:
    Still faithful to their respective answer options,
    More concretely testable from the video evidence,
    Better aligned with the available context.\\
Follow these reasoning steps:\\
1. Analyze Verification Feedback\\
    Identify why the previous verification failed (e.g., missing actions, unclear subject, no visible timing, or too abstract).
    Determine what type of visual or contextual evidence is available or missing.\\
2. Regenerate Option-based Hypotheses\\
    For each option, rewrite its hypothesis to make it more specific, testable, and observable within the context.
    Preserve each option’s core meaning (do not alter its logical claim).
    Avoid unverifiable statements (e.g., emotions, intentions, unseen causes).
    If an option is clearly impossible to verify from context, 
    note that no valid hypothesis can be formed.\\
    
[Inputs]\\
Question: \{question\}\\
Options: \{option\}\\
Previous hypotheses: \{previous\_hypotheses\}\\
Verification feedback: \{verification\_feedback\}\\
Context summary: \{video summary\}\\

[Output format]\\
Return option (It must be a complete option, not just numbers) and hypothesis.
\end{tcolorbox}